%% file: MICCAI2026-main_conference_paper_template.tex
\documentclass[runningheads]{llncs}
\usepackage[T1]{fontenc}
\usepackage{graphicx,verbatim}
\usepackage{caption}
\usepackage{pgfplots}
\usepackage{booktabs}
\usepackage{subcaption}
\usepackage{siunitx}
\usepackage{caption}
\usepackage{amsfonts}
\usepackage{amsmath}
\usepackage{booktabs}
\usepackage{multirow}
\usepackage[table]{xcolor} 
\usepackage{enumitem}
\newcommand{\nbold}[1]{\noindent\textbf{#1}}

\begin{document}
\title{WaveDiT: Distribution-Aware Wavelet Flow Matching for Efficient 3D Brain MRI Synthesis}

\titlerunning{WaveDiT}
%
\author{Danilo Danese\inst{1} \and
Angela Lombardi \inst{1} \and Giuseppe Fasano\inst{1} \and Matteo Attimonelli \inst{1,2} \and Tommaso Di Noia\inst{1}}
\authorrunning{D. Danese et al.}
%
\institute{Politecnico di Bari, Italy
\email{\{name.surname\}@poliba.it}\\ \and
Sapienza University of Rome, Italy }


  
\maketitle              
\begin{abstract}
Large and demographically balanced datasets are essential for reliable neuroimaging biomarkers. Full-resolution 3D brain MRI synthesis can support data augmentation in this setting, but existing approaches either incur prohibitive computational cost at volumetric scale or rely on lossy latent compression that may compromise anatomical detail. As a result, practical 3D generative augmentation often requires specialized compute infrastructure.
We propose WaveDiT, a conditional flow matching framework operating in the coefficient space of a 3D Haar Discrete Wavelet Transform. The model combines factorized spatio-depth attention with band-wise heteroscedastic uncertainty modeling derived from higher-order wavelet statistics. Predicted log-variance is integrated directly into both the flow objective and conditioning pathway, enabling adaptive precision consistent with the heavy-tailed and input-dependent variance structure of anatomical detail. This formulation supports full-resolution 3D synthesis under practical memory and time constraints on a single modern GPU.
Evaluation on a multi-site cohort demonstrates improved alignment between generated and real MRI distributions, together with enhanced downstream brain age prediction and region-level anatomical agreement relative to diffusion, latent, and wavelet-based baselines. Code is available at \url{https://github.com/sisinflab/WaveDiT}.


\end{abstract}

\input{SECTIONS/1_introduction}
\input{SECTIONS/2_methodology}
\input{SECTIONS/3_results}
\input{SECTIONS/4_conclusion}

\subsubsection{\ackname} This work was partially supported by the following projects: “LIFE: the itaLian system wIde Frailty nEtwork”; We acknowledge the CINECA award under the ISCRA initiative (Projects IsCc1 SynBrain, IsCd1 FlowMRI, IsCd3 EMBRAIN) for the availability of high performance computing resources and support”; This work has been carried out while \textit{Matteo Attimonelli} was enrolled in the Italian National Doctorate on Artificial Intelligence run by Sapienza University of Rome in collaboration with \textit{Politecnico Di Bari}.

\bibliographystyle{splncs04}
\bibliography{bibliography}

\end{document}

%% file: SECTIONS/1_introduction.tex
\section{Introduction}

High-resolution brain MRI is fundamental to neuroimaging research, supporting clinically relevant tasks such as brain age prediction (BAP), disease-risk stratification, and longitudinal monitoring of neurodegeneration~\cite{cole2017predicting,rahman2025understanding}. Developing robust biomarkers for these tasks typically requires large, demographically balanced cohorts~\cite{benkarim2022population}. However, acquiring such datasets is challenging due to high acquisition costs, privacy constraints, and site-specific heterogeneity. This scarcity is particularly pronounced in specific age ranges, where sparsity can induce biased regression and inflated uncertainty in normative aging trajectories~\cite{dinsdale2021learning}. Generative modeling offers a promising path forward, enabling the synthesis of balanced cohorts to augment scarce clinical data~\cite{Chintapalli2024-hy}.
Despite recent progress, scaling generative models to full-resolution 3D MRI remains computationally demanding. Pixel-space diffusion models achieve high fidelity but require hundreds to thousands of iterative denoising steps over millions of voxels~\cite{DBLP:conf/nips/HoJA20}, resulting in substantial training and inference costs. Training such models at full volumetric resolution often necessitates high-memory GPUs and prolonged compute schedules, limiting accessibility outside specialized infrastructure. Latent diffusion reduces computational demand through learned compression~\cite{DBLP:conf/cvpr/RombachBLEO22,pinaya2022brain}; however, this compression is inherently lossy and may discard fine-grained anatomical detail or introduce reconstruction artifacts, particularly in regions with subtle cortical structure~\cite{Muller-Franzes2023-ju}.
The Discrete Wavelet Transform (DWT) provides an alternative representation that preserves invertibility while reducing spatial dimensionality. By decomposing volumes into low-frequency approximation and high-frequency subbands, wavelets maintain anatomical structure without learned compression artifacts. Recent wavelet-based generative approaches demonstrate that operating in the wavelet domain is effective for 3D MRI synthesis~\cite{DBLP:conf/miccai/FriedrichWBDC24,danese2025flowlet}. However, existing methods typically treat all subbands uniformly, applying band-agnostic objectives and conditioning strategies. In practice, wavelet subbands exhibit markedly different statistical properties: approximation coefficients remain near-Gaussian, whereas high-frequency bands are sparse, heavy-tailed, and strongly heteroscedastic, with distributions that evolve along the generative trajectory. This heteroscedastic structure implies input-dependent predictive uncertainty, making uniform loss weighting suboptimal. Heteroscedastic uncertainty modeling has improved robustness in regression~\cite{DBLP:conf/nips/KendallG17,DBLP:conf/iclr/SeitzerTAM22} and medical image registration~\cite{DBLP:conf/miccai/ZhangPALYSSWD24}, and its integration within flow-based generative modeling remains limited.
To address these challenges, we propose (i)~\textbf{WaveDiT}, a conditional flow matching framework that operates directly in the wavelet domain. WaveDiT extends the Hourglass Diffusion Transformer (HDiT)~\cite{DBLP:conf/icml/CrowsonBBAKS24} to volumes with factorized intra-slice and inter-slice attention, avoiding computational complexity over full 3D self-attention in the wavelet domain. Furthermore, we introduce (ii)~\textbf{Morpheus}, a state-aware auxiliary network that predicts per-band precision from higher-order wavelet statistics. Morpheus enables a Bayesian heteroscedastic loss and frequency-aware conditioning that adapts to the signal's complexity along the flow trajectory.
We evaluate the proposed framework using a multi-level protocol that combines global distributional metrics with downstream brain age prediction and region-level anatomical analysis.

%% file: SECTIONS/2_methodology.tex
\section{Methods}

\input{TABLES/heteroscedastic_analysis}
\subsection{Wavelet modeling}
WaveDiT operates in the coefficient space of a single-level 3D Haar DWT, 
which decomposes each volume as
$
\mathcal{W}: \mathbb{R}^{1 \times D \times H \times W} \rightarrow 
\mathbb{R}^{8 \times D' \times H' \times W'},
$
with $D' = D/2$, $H' = H/2$, $W' = W/2$, producing one low-frequency 
approximation subband ($LLL$) and seven directional HF detail subbands 
($LLH, LHL, \dots, HHH$).

Analysis of the training volumes reveals two key statistical properties. First, the signal energy concentrates heavily in the approximation band: $LLL$ contains 98.11\% of total energy. Second, wavelet coefficient distributions change drastically along the flow trajectory (Table~\ref{tab:merged_stats}). At $t=0$ (pure noise), all bands exhibit Gaussian statistics with kurtosis $\kappa \approx 3$. As the flow progresses toward $t=1$ (data), the $LLL$ band remains near-Gaussian ($\kappa \approx 5$), but HF bands diverge sharply: single-axis subbands develop $\kappa \in [27, 31]$, two-axis subbands reach $\kappa \in [84, 93]$, and the isotropic $HHH$ band peaks at $\kappa \approx 270$ a ratio of $89.8\times$ relative to $t=0$. This evolving distributional contrast, confirmed by Jarque-Bera tests that reject normality at $t=1$ with $p < 10^{-10}$, makes kurtosis a natural signal-noise discriminator. Beyond distributional shape, HF coefficients are also strongly heteroscedastic. Their local variance varies by roughly eight orders of magnitude across space, increasing near tissue boundaries (steep gradients) and decreasing in homogeneous regions. This variance is input-dependent, with diminishing correlation as frequency increases.

\subsection{Morpheus: State-Aware Uncertainty Modelling}
\label{sec:morpheus}

The heteroscedastic, heavy-tailed statistics have direct implications for optimization. Standard flow matching employs uniform MSE weighting, treating all wavelet bands and spatial locations identically. Such fixed-precision losses over-penalize errors at anatomical boundaries (high variance) and under-penalize homogeneous regions (low variance). WaveDiT addresses this mismatch with {Morpheus}, a lightweight network that predicts input-dependent band-wise uncertainty and influences both optimization and generation.

\nbold{Feature Extraction.} Unlike conventional schedulers that condition solely on timestep $t$, Morpheus computes the statistical signature of the current noisy state $\mathbf{x}_t$ of an input $\mathbf{x}$ at time $t$ along the flow trajectory.
For each channel $c$, we extract six statistics:
\newline\newline\newline
\begin{itemize}
\small
    \item \textbf{Mean} $\mu_c$ and \textbf{Standard Deviation} $\sigma_c$: First and second moments;
    \item \textbf{Maximum Absolute Value} $\max|x_c|$: Outlier detection;
    \item \textbf{L2 Norm} $\|x_c\|_2 / \sqrt{N}$: Normalized energy per band;
    \item \textbf{Skewness} $\gamma_1 = \mathbb{E}[(x - \mu)^3] / \sigma^3$: Distribution asymmetry;
    \item \textbf{Kurtosis} $\gamma_2 = \mathbb{E}[(x - \mu)^4] / \sigma^4$: Tail heaviness.
\end{itemize}
These features are concatenated with a sinusoidal time embedding and processed by an MLP (parameters $\psi$), 
to produce the band-wise log-variance $\mathbf{s}_\psi(\mathbf{x}_t, t)$.

\nbold{Bayesian Heteroscedastic Objective.}
The generative trajectory follows Rectified Conditional Flow Matching~\cite{DBLP:conf/iclr/LiuG023}. Given $\mathbf{x}_1 \sim p_{\text{data}}$ 
and $\mathbf{x}_0 \sim \mathcal{N}(0,I)$, the interpolation $\mathbf{x}_t = (1-t)\mathbf{x}_0 + t\mathbf{x}_1$ defines a linear path with target velocity $\mathbf{v}_{\text{target}} = \frac{\mathbf{x}_1 - \mathbf{x}_t}{1-t+\epsilon}$, with small $\epsilon > 0$ preventing divergence as 
$t \to 1$. Following heteroscedastic regression~\cite{DBLP:conf/nips/KendallG17,DBLP:conf/iclr/SeitzerTAM22}, 
velocity prediction is modeled as $p(\mathbf{v} \mid \mathbf{v}_\theta, \mathbf{s}) = \mathcal{N}(\mathbf{v}; \mathbf{v}_\theta, e^{\mathbf{s}})$, yielding the objective (backbone $\theta$, Morpheus $\psi$):
\begin{equation}
\small
\mathcal{L} = \mathbb{E}_{\mathbf{x}_0, \mathbf{x}_1, t}
\left[
\frac{1}{2} e^{-\mathbf{s}_\psi(\mathbf{x}_t,t)}
\|\mathbf{v}_\theta(\mathbf{x}_t,t,cond) - \mathbf{v}_{\text{target}}\|^2
+ \frac{1}{2}\mathbf{s}_\psi(\mathbf{x}_t,t)
\right].
\label{eq:bayesian_loss}
\end{equation}
Here, $e^{-\mathbf{s}}$ adaptively reweights the velocity loss during training, down-weighting inherently unpredictable high-frequency content, while $\frac{1}{2}\mathbf{s}$ prevents trivial variance inflation. This results in state-dependent precision, with higher predictive variance during early noisy states and progressively sharper weighting as structured anatomy emerges.

\nbold{Frequency Conditioning.} Morpheus also conditions the backbone. The predicted log-variances $\mathbf{s}$ are linearly projected and concatenated with time, slice, and metadata embeddings to form a \emph{frequency hint}. Active during both training and sampling, this pathway allows the backbone to adapt its predictions to the 
current reliability of each wavelet band.

\begin{figure*}[t]
\centering
\includegraphics[width=\textwidth]{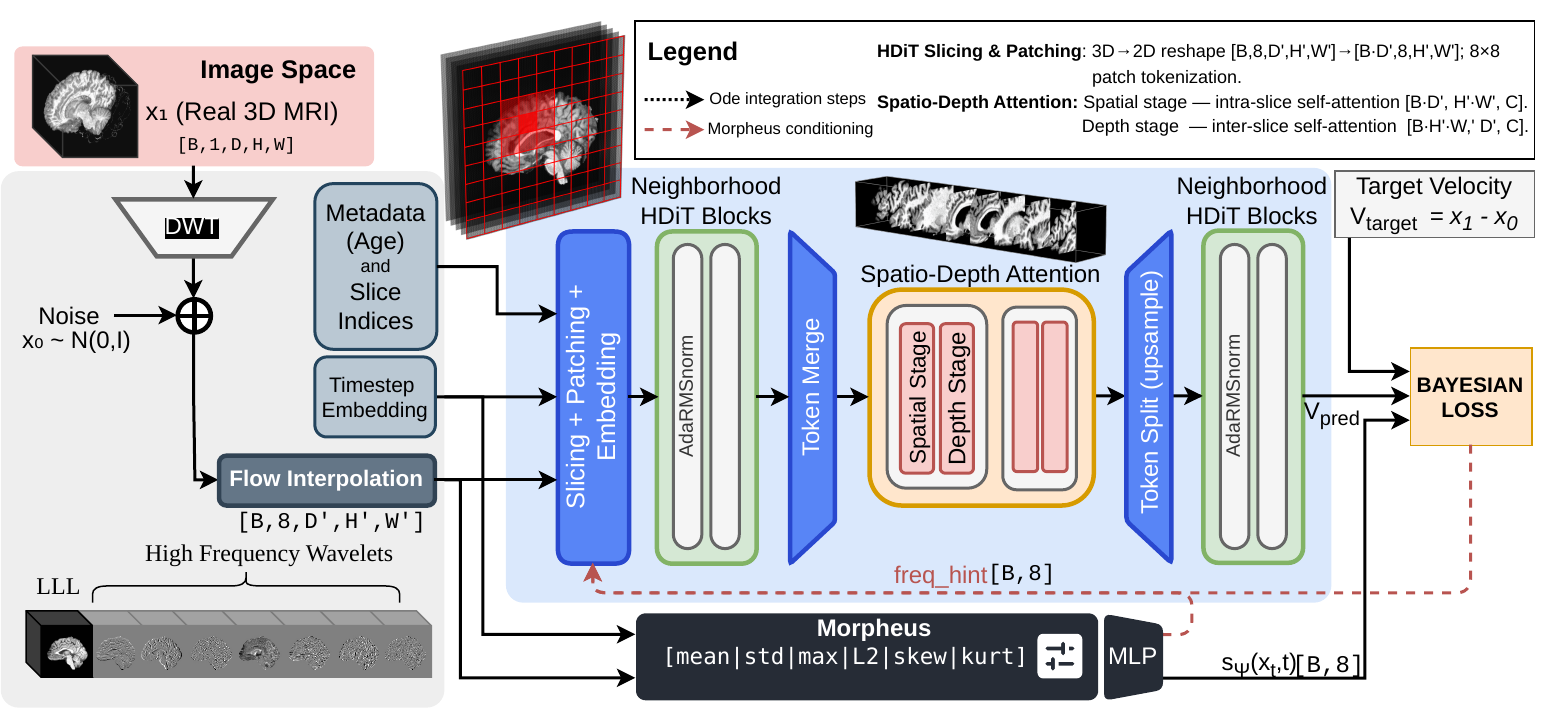}
\caption{Training pipeline: wavelet decomposition, HDiT backbone with Morpheus scheduling, and Bayesian heteroscedastic loss. }
\label{fig:architecture}
\end{figure*}

\subsection{The WaveDiT Architecture}
Even in the reduced wavelet domain, global self-attention over the full 3D coefficient tensor remains prohibitive. For a $224^3$ input, the DWT yields $D' \!=\! H' \!=\! W' \!=\! 112$, producing $N = D'H'W' \approx 1.4 \times 10^6$ spatial tokens. Full 3D attention scales as $\mathcal{O}(N^2) \approx 2 \times 10^{12}$. WaveDiT avoids this extending the hourglass design~\cite{DBLP:conf/icml/CrowsonBBAKS24} to 3D, treating the volume as a batch of 2D slices and restores volumetric coherence through factorized depth attention, reducing complexity to $\mathcal{O}(D'(H'W')^2 + H'W'D'^2) \approx 1.8 \times 10^{10}$, a $\sim$110$\times$ reduction.

\nbold{Slice Processing.}
The wavelet tensor $\mathbf{w} \in \mathbb{R}^{B \times 8 \times D' \times H' \times W'}$ is reshaped into a batch of $B\!\cdot\!D'$ 2D slices and tokenized with 2D patch embeddings. To recover explicit slice position lost in this operation, we encode each slice with fixed random Fourier features, which are combined with sinusoidal time embedding, demographic metadata (age), and the Morpheus frequency hint into a global conditioning vector. All transformer blocks are modulated via AdaRMSNorm~\cite{DBLP:conf/nips/ZhangS19a} as a multiplicative scale, injecting time-, slice-, and frequency-dependent conditioning without concatenating condition vectors to spatial tokens.

\nbold{Level 1: Neighborhood Attention.}
At the highest spatial resolution, shallow transformer layers employ 2D sliding-window attention. Each query token attends only to keys within a local $K \!\times\! K$ neighborhood, reducing per-layer complexity from $\mathcal{O}((H'W')^2)$ to $\mathcal{O}(H'W'K^2)$. This locality inductive bias aligns naturally with HF coefficients, which encode edges and tissue boundaries with strong local spatial correlation. AxialRoPE~\cite{DBLP:conf/eccv/HeoPHY24} is applied to queries and keys, encoding relative 2D positions while preserving translation equivariance.

\nbold{Level 2: Factorized Spatio-Depth Attention.}
Deeper layers in the model hierarchy employ our two-stage factorized attention block, which restores volumetric consistency without incurring the cost of full 3D self-attention. In the spatial (intra-slice) stage, we apply global self-attention independently within each 2D slice to capture long-range in-plane dependencies such as bilateral symmetry and ventricle-to-cortex relationships. In the subsequent depth (inter-slice) stage, tokens at each spatial location $(h,w)$ are grouped and attend along the depth axis, enabling cross-slice propagation of anatomical context (e.g., continuity of cortical sulci or the corpus callosum along the Z-axis). This mechanism preserves intra-slice structure and inter-slice consistency at reduced cost.

\subsection{Inference}
Noise $\mathbf{z}_0 \sim \mathcal{N}(0,I)$ is integrated from $t=0$ to $t=1$  via a second-order Heun solver, with Morpheus providing frequency conditioning at each step. The inverse 3D Haar DWT reconstructs the synthesized volume in voxel space.

%% file: TABLES/heteroscedastic_analysis.tex
\begin{table}[t]
\centering
\caption{Wavelet band statistics and kurtosis evolution along the flow trajectory.
\textbf{Left:} per-band kurtosis at $t=1$ and Pearson correlation between local HF variance and LLL intensity.
\textbf{Right:} subband kurtosis as a function of timesteps.}
\label{tab:merged_stats}
\setlength{\tabcolsep}{1.5mm} 
\renewcommand{\arraystretch}{1.1} 
\small

\begin{minipage}[c]{0.57\textwidth}
\begin{tabular}{l r@{\,$\pm$\,}l c | r@{\,$\pm$\,}l}
\toprule
Band & \multicolumn{2}{c}{$\kappa$ at $t=1$} & Ratio & \multicolumn{2}{c}{Pearson $r$} \\
\midrule
LLL &   5.06 & 0.87 & $1.7\times$ & \multicolumn{2}{c}{---} \\
LLH &  29.77 & 2.71 & $9.9\times$ & 0.559 & 0.037 \\
LHL &  30.99 & 3.72 & $10.3\times$ & 0.592 & 0.040 \\
HLL &  27.32 & 2.88 & $9.1\times$ & 0.466 & 0.092 \\
LHH &  83.80 & 17.68 & $27.9\times$ & 0.431 & 0.044 \\
HLH &  87.95 & 18.17 & $29.3\times$ & 0.354 & 0.068 \\
HHL &  92.53 & 15.36 & $30.8\times$ & 0.399 & 0.078 \\
HHH & $\mathbf{269.44}$ & $\mathbf{50.66}$ & $\mathbf{89.8\times}$ & 0.200 & 0.066 \\
\bottomrule
\end{tabular}
\end{minipage}\hfill
\begin{minipage}[c]{0.41\textwidth}
\definecolor{colorLLL}{HTML}{435065}
\definecolor{colorLLH}{HTML}{435065}
\definecolor{colorLHL}{HTML}{435065}
\definecolor{colorHLL}{HTML}{435065}
\definecolor{colorLHH}{HTML}{23b3c9}
\definecolor{colorHLH}{HTML}{23b3c9}
\definecolor{colorHHL}{HTML}{23b3c9}
\definecolor{colorHHH}{HTML}{f79430}
\begin{tikzpicture}
\begin{axis}[
    width=\linewidth,
    height=5.2cm,
    xlabel={Timestep},
    ylabel={Kurtosis},
    xlabel style={yshift=2ex},
    ylabel style={yshift=-4ex},
    xmin=0.9, xmax=1.0,
    legend pos=north west,
    legend style={draw=none, fill=none, font=\tiny, row sep=-2pt},
    grid=major,
    cycle list name=color list,
    tick label style={font=\tiny},
    label style={font=\scriptsize},
]
\addplot[fill=colorLLL, fill opacity=0.2, draw=none, forget plot] coordinates {(0.9091,4.1625) (0.9192,4.1685) (0.9293,4.1736) (0.9394,4.1779) (0.9495,4.1815) (0.9596,4.1843) (0.9697,4.1865) (0.9798,4.1880) (0.9899,4.1888) (1.0000,4.1891) (1.0000,5.9276) (0.9899,5.9265) (0.9798,5.9232) (0.9697,5.9176) (0.9596,5.9095) (0.9495,5.8989) (0.9394,5.8855) (0.9293,5.8693) (0.9192,5.8502) (0.9091,5.8280)};
\addplot[color=colorLLL, solid, thick] coordinates {(0.9091,4.9953) (0.9192,5.0093) (0.9293,5.0215) (0.9394,5.0317) (0.9495,5.0402) (0.9596,5.0469) (0.9697,5.0521) (0.9798,5.0556) (0.9899,5.0577) (1.0000,5.0584)};
\addlegendentry{LLL}
\addplot[fill=colorLLH, fill opacity=0.2, draw=none, forget plot] coordinates {(0.9091,5.5518) (0.9192,6.5713) (0.9293,8.0133) (0.9394,10.0145) (0.9495,12.6894) (0.9596,16.0396) (0.9697,19.8190) (0.9798,23.4432) (0.9899,26.0916) (1.0000,27.0567) (1.0000,32.4775) (0.9899,31.2434) (0.9798,27.9050) (0.9697,23.4473) (0.9596,18.9021) (0.9495,14.9209) (0.9394,11.7399) (0.9293,9.3361) (0.9192,7.5781) (0.9091,6.3146)};
\addplot[color=colorLLH, dashed, thick] coordinates {(0.9091,5.9332) (0.9192,7.0747) (0.9293,8.6747) (0.9394,10.8772) (0.9495,13.8052) (0.9596,17.4709) (0.9697,21.6331) (0.9798,25.6741) (0.9899,28.6675) (1.0000,29.7671)};
\addlegendentry{LLH}
\addplot[fill=colorLHL, fill opacity=0.2, draw=none, forget plot] coordinates {(0.9091,5.3797) (0.9192,6.3591) (0.9293,7.7619) (0.9394,9.7357) (0.9495,12.4119) (0.9596,15.8110) (0.9697,19.6976) (0.9798,23.4707) (0.9899,26.2554) (1.0000,27.2757) (1.0000,34.7075) (0.9899,33.2391) (0.9798,29.3119) (0.9697,24.1911) (0.9596,19.1305) (0.9495,14.8448) (0.9394,11.5276) (0.9293,9.0886) (0.9192,7.3444) (0.9091,6.1132)};
\addplot[color=colorLHL, dotted, thick] coordinates {(0.9091,5.7465) (0.9192,6.8518) (0.9293,8.4253) (0.9394,10.6317) (0.9495,13.6283) (0.9596,17.4708) (0.9697,21.9444) (0.9798,26.3913) (0.9899,29.7472) (1.0000,30.9916)};
\addlegendentry{LHL}
\addplot[fill=colorHLL, fill opacity=0.2, draw=none, forget plot] coordinates {(0.9091,4.9056) (0.9192,5.7105) (0.9293,6.8799) (0.9394,8.5551) (0.9495,10.8774) (0.9596,13.9006) (0.9697,17.4352) (0.9798,20.9159) (0.9899,23.4977) (1.0000,24.4428) (1.0000,30.2016) (0.9899,28.9268) (0.9798,25.5567) (0.9697,21.2053) (0.9596,16.9219) (0.9495,13.2799) (0.9394,10.4375) (0.9293,8.3301) (0.9192,6.8132) (0.9091,5.7375)};
\addplot[color=colorHLL, dashdotted, thick] coordinates {(0.9091,5.3215) (0.9192,6.2619) (0.9293,7.6050) (0.9394,9.4963) (0.9495,12.0787) (0.9596,15.4113) (0.9697,19.3202) (0.9798,23.2363) (0.9899,26.2122) (1.0000,27.3222)};
\addlegendentry{HLL}
\addplot[fill=colorLHH, fill opacity=0.2, draw=none, forget plot] coordinates {(0.9091,3.1530) (0.9192,3.2504) (0.9293,3.4304) (0.9394,3.7861) (0.9495,4.5474) (0.9596,6.3325) (0.9697,10.9091) (0.9798,22.9622) (0.9899,47.9724) (1.0000,66.1193) (1.0000,101.4723) (0.9899,71.4709) (0.9798,33.9371) (0.9697,15.9631) (0.9596,8.7351) (0.9495,5.7557) (0.9394,4.4331) (0.9293,3.7978) (0.9192,3.4701) (0.9091,3.2904)};
\addplot[color=colorLHH, dashed, thick] coordinates {(0.9091,3.2217) (0.9192,3.3603) (0.9293,3.6141) (0.9394,4.1096) (0.9495,5.1515) (0.9596,7.5338) (0.9697,13.4361) (0.9798,28.4497) (0.9899,59.7217) (1.0000,83.7958)};
\addlegendentry{LHH}
\addplot[fill=colorHLH, fill opacity=0.2, draw=none, forget plot] coordinates {(0.9091,3.1081) (0.9192,3.1768) (0.9293,3.3049) (0.9394,3.5621) (0.9495,4.1256) (0.9596,5.4973) (0.9697,9.2376) (0.9798,20.1853) (0.9899,47.0475) (1.0000,69.7795) (1.0000,106.1159) (0.9899,75.4909) (0.9798,36.2384) (0.9697,16.9388) (0.9596,9.1452) (0.9495,5.9431) (0.9394,4.5270) (0.9293,3.8487) (0.9192,3.4996) (0.9091,3.3085)};
\addplot[color=colorHLH, dotted, thick] coordinates {(0.9091,3.2083) (0.9192,3.3382) (0.9293,3.5768) (0.9394,4.0445) (0.9495,5.0343) (0.9596,7.3213) (0.9697,13.0882) (0.9798,28.2118) (0.9899,61.2692) (1.0000,87.9477)};
\addlegendentry{HLH}
\addplot[fill=colorHHL, fill opacity=0.2, draw=none, forget plot] coordinates {(0.9091,3.1309) (0.9192,3.2146) (0.9293,3.3702) (0.9394,3.6809) (0.9495,4.3567) (0.9596,5.9842) (0.9697,10.3580) (0.9798,22.9159) (0.9899,52.8208) (1.0000,77.1711) (1.0000,107.8883) (0.9899,73.7289) (0.9798,33.5258) (0.9697,15.3195) (0.9596,8.3137) (0.9495,5.5133) (0.9394,4.2948) (0.9293,3.7169) (0.9192,3.4212) (0.9091,3.2600)};
\addplot[color=colorHHL, dashdotted, thick] coordinates {(0.9091,3.1955) (0.9192,3.3179) (0.9293,3.5436) (0.9394,3.9879) (0.9495,4.9350) (0.9596,7.1490) (0.9697,12.8387) (0.9798,28.2208) (0.9899,63.2749) (1.0000,92.5297)};
\addlegendentry{HHL}
\addplot[fill=colorHHH, fill opacity=0.2, draw=none, forget plot] coordinates {(0.9091,3.0547) (0.9192,3.0914) (0.9293,3.1615) (0.9394,3.3071) (0.9495,3.6453) (0.9596,4.5567) (0.9697,7.5708) (0.9798,20.6842) (0.9899,90.5887) (1.0000,218.7834) (1.0000,320.0975) (0.9899,128.4823) (0.9798,33.7352) (0.9697,12.0023) (0.9596,6.2679) (0.9495,4.4022) (0.9394,3.6811) (0.9293,3.3632) (0.9192,3.2082) (0.9091,3.1265)};
\addplot[color=colorHHH, dotted, thick] coordinates {(0.9091,3.0906) (0.9192,3.1498) (0.9293,3.2624) (0.9394,3.4941) (0.9495,4.0238) (0.9596,5.4123) (0.9697,9.7866) (0.9798,27.2097) (0.9899,109.5355) (1.0000,269.4405)};
\addlegendentry{HHH}
\end{axis}
\end{tikzpicture}
\end{minipage}
\end{table}

%% file: SECTIONS/3_results.tex
\newpage
\section{Experiments and Results}
\label{sec:experiments}

\begin{figure}[t]
\centering
\includegraphics[width=\textwidth]{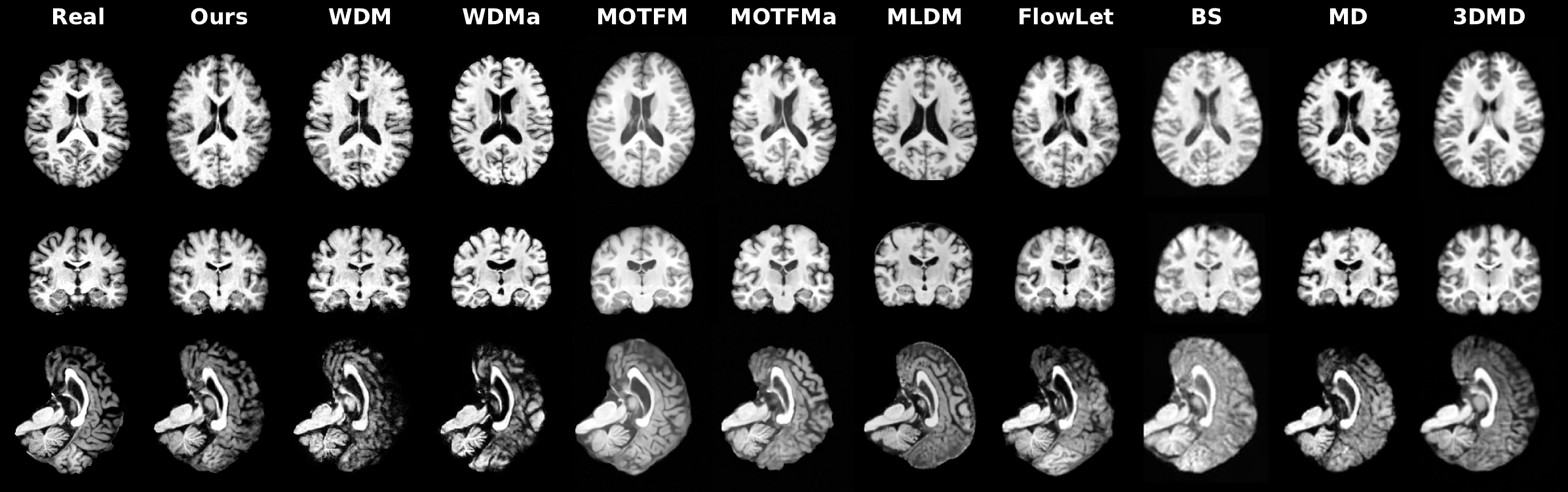}
\caption{Visual comparison of models. Axial, coronal, and sagittal views of a real 72 years old subject and age-conditioned generations at the same target age.}
\label{fig:qualitative}
\end{figure}

\nbold{Dataset.}
We follow the multi-site cohort introduced in~\cite{danese2025flowlet}, merging OpenBHB~\cite{openbhb}, ADNI~\cite{adni}, and OASIS-3~\cite{oasis} ($5{,}989$ cognitively normal subjects, ages 5.9-95.5\,yrs) to mitigate the elderly under-representation inherent in OpenBHB alone. All volumes undergo standard preprocessing: bias-field correction~\cite{Tustison2010-sx}, affine registration to MNI152~\cite{JENKINSON2002825,FONOV2009S102}, skull stripping~\cite{Smith2002-lm}, and $[0,1]$ normalization, yielding an isotropic $182\!\times\!218\!\times\!182$ resolution. A subject-disjoint 20\% hold-out is reserved exclusively for BAP evaluation, while all generative models are trained on the last 80\% following strict subject-level separation to prevent identity leakage.

\nbold{Implementation.}
WaveDiT (patch $8\!\times\!8$, two HDiT levels: depth 2 each, width 1024, FFN dim 4096) uses level 1 neighborhood attention ($d_{\text{head}}=64$, window $K=7$) and level 2 spatio-depth attention ($d_{\text{head}}=64$), totalising $\sim$142M parameters. The Morpheus MLP has two hidden layers (width 128) with SiLU activations. The model is trained with AdamW (lr=$10^{-4}$, weight decay $10^{-2}$, bs=4, dropout 0.1) for 200 epochs on a single H100 ($\sim$12GB VRAM at bs=1). Checkpoints are selected by best validation loss; Morpheus is jointly optimized.
Training completes in 26 hours on a single H100, compared to $\sim$6 days for FlowLet and $\sim$7 days for WDM under the same hardware setting. At inference, WaveDiT generates a full 3D volume in $\sim$1 second using 10 steps, compared to $\sim$6 seconds for 10 steps in FlowLet and $\sim$150 seconds for 1000 steps in WDM.

\nbold{Baselines.}
We compare nine baselines spanning diffusion, latent diffusion, and flow-based paradigms: {WDM}~\cite{DBLP:conf/miccai/FriedrichWBDC24}, {MD}~\cite{DBLP:journals/corr/abs-2211-03364}, {3DMD}~\cite{DBLP:journals/tmi/WangLSWSC25}, {MLDM}~\cite{pinaya2022brain}, {BS}~\cite{DBLP:journals/natmi/TudosiuPCDPBFGGNOC24}, {MOTFM}~\cite{yazdani2025flow}, and {FlowLet}~\cite{danese2025flowlet}; the latter also provides age-conditioned variants of WDM and MOTFM, denoted {WDMa} and {MOTFMa}.
All models are trained on the same generative training cohort. For evaluation, each model generated 3000 synthetic volumes; conditional models spanned the training age range linearly, unconditional samples were assigned ages from the training distribution.

\nbold{Evaluation protocol.}
Quantitative global metrics in volumetric MRI can be skewed by the large proportion of empty background voxels, potentially hiding anatomical inconsistencies; therefore, we follow a three-level evaluation protocol proposed in~\cite{danese2025flowlet}.
\emph{(i)~Global distributional metrics}: FID and MMD quantify distributional alignment between generated and real MRIs using features extracted from a medical-pretrained ResNet-50~\cite{DBLP:journals/corr/abs-1904-00625}. MS-SSIM is computed intra-set as the average pairwise similarity among generated samples, where lower values indicate greater sample diversity.
\emph{(ii)~Brain Age Prediction (BAP)}: following~\cite{de2024explainable}, we train a separate 3D DenseNet for each generative method using its synthetic samples as data augmentation to predict chronological age from MRI. Each BAP model is then evaluated on the same held-out set of real subjects older than 44 years, and performance is reported as Mean Absolute Error in years. This age range is under-represented in OpenBHB, where data sparsity can affect prediction accuracy~\cite{dinsdale2021learning}. \emph{(iii)~Region-of-interest (ROI) analysis}: each synthetic volume and an age-matched real counterpart are independently segmented~\cite{DBLP:journals/neuroimage/HenschelCEDFR20} into 95 cortical and subcortical regions.
Region-wise intensity MAE, KL divergence, and Dice coefficient are computed over the union of real and synthetic segmentations and averaged across all ROIs, following the procedures of~\cite{danese2025flowlet}.
These protocols jointly ensure that evaluation captures age-specific anatomical fidelity that is robust to both distributional overfitting and diversity collapse.

\input{TABLES/Table_global_metrics}

\nbold{Generative quality.}
Table~\ref{tab:gen_quality} reports global distributional metrics, with visual examples in Figure~\ref{fig:qualitative}. 
At 10 steps, WaveDiT-CFM achieves the lowest metrics among all methods, improving over the conditioned wavelet-based baseline FlowLet.
Compared to WDM, which also operates in the wavelet domain but requires 1000 diffusion steps, WaveDiT-CFM achieves lower FID, indicating that the proposed flow-based formulation is well suited to high-resolution wavelet coefficients. 
Across internal objectives, CFM outperforms RFM~\cite{DBLP:conf/iclr/LipmanCBNL23} and OTFM~\cite{DBLP:conf/iclr/ChenL24}, suggesting that conditional trajectories yield a more favorable balance between fidelity and efficiency than constant-velocity or optimal-transport in this setting.

\input{TABLES/table_BAP_ROI}

\nbold{Downstream evaluation.}
Table~\ref{tab:clinical} summarizes BAP and ROI metrics. 
WaveDiT-CFM achieves the lowest BAP MAE, 
outperforming the reference model trained solely on real data under the same protocol

and other conditional baselines when used for synthetic augmentation. 
It also achieves the best ROI-level scores, with lower iMAE and KL divergence and higher Dice than alternatives.
Notably, WDM and MD achieve competitive FID while exhibiting reduced Dice and higher KLD, reinforcing that global metrics may underestimate discrepancies in regional brain structure and motivating the use of BAP and ROI-level analysis as complementary endpoints.

\nbold{Ablations.}
The Morpheus module has a substantial impact on both global and downstream metrics. 
Removing Morpheus degrades FID and MMD and worsens BAP and ROI scores, indicating that uniform MSE is insufficient under the statistics of high-frequency wavelet bands, confirming the importance of state-aware precision control for stable training and improved anatomical fidelity. 
Among flow objectives, CFM provides the strongest overall trade-off: RFM exhibits degraded global and ROI performance, while OTFM improves MS-SSIM but does not match CFM on FID or anatomical scores, suggesting that CFM trajectories better align with wavelet-domain statistics under a limited step budget. 
Finally, the voxel-space variant, although similar in FID and MMD, yields higher BAP error and degraded ROI scores, indicating that matching global scores in voxel space does not guarantee preservation of clinically relevant anatomy.

%% file: TABLES/Table_global_metrics.tex
\begin{table*}[t]
\caption{Generative quality. Best in \textbf{bold}, second \underline{underlined}.
$^\dagger$\!Unconditional model.
All pairwise comparisons against WaveDiT-CFM 10 steps are statistically significant (Wilcoxon rank-sum, Bonferroni-corrected $p<0.001$).
Metrics are computed over 10 bootstrap resamples of 500 generated samples.
Standard deviations ($\leq 10^{-3}$) are omitted for conciseness.}
\label{tab:gen_quality}
\centering
\begin{tabular}{@{}p{0.48\textwidth}@{\hspace{0.04\textwidth}}p{0.48\textwidth}@{}}
\begin{minipage}[t]{\linewidth}
\centering
\subcaption{External baselines}
\label{tab:baselines}
\resizebox{\linewidth}{!}{%
\small
\begin{tabular}{@{}cl c c c c@{}}
\toprule
& {Method} & {Steps} & {FID\,$\downarrow$} & {MMD\,$\downarrow$} & {MS-SSIM\,$\downarrow$} \\
\midrule
\multirow{9}{*}{\rotatebox[origin=c]{90}{\scriptsize\textit{Baselines}}}
& 3DMD$^\dagger$    & 1000 & 0.0307  & 0.00060 & 0.887 \\
& MD$^\dagger$      & 1000 & 0.0113  & 0.00041 & 0.898 \\
& WDM$^\dagger$     & 1000 & 0.0045  & 0.00011 & 0.908 \\
& WDMa              & 1000 & \underline{0.0044}  & 0.00014 & 0.900 \\
& MOTFM$^\dagger$   &   50 & 0.1154  & 0.00455 & 0.918 \\
& MOTFMa            &   50 & 0.1124  & 0.00433 & 0.848 \\
& MLDM              &  100 & 0.0704  & 0.00272 & 0.884 \\
& BS                &   -- & 0.0477  & 0.00178 & 0.852 \\
& FlowLet           &   10 & \multicolumn{1}{c}{0.0117} & \multicolumn{1}{c}{0.00040} & \multicolumn{1}{c}{0.869} \\
\bottomrule
\end{tabular}%
}
\end{minipage}
&
\begin{minipage}[t]{\linewidth}
\centering
\subcaption{WaveDiT (ours) \& ablations}
\label{tab:ours_ablations}
\resizebox{\linewidth}{!}{%
\small
\begin{tabular}{@{}cl c c c c@{}}
\toprule
& {Method} & {Steps} & {FID\,$\downarrow$} & {MMD\,$\downarrow$} & {MS-SSIM\,$\downarrow$} \\
\midrule
\multirow{3}{*}{\rotatebox[origin=c]{90}{\scriptsize\textit{Ours}}}
& WaveDiT-RFM       &   10 & 0.1811  & 0.00719 & 0.862 \\
& WaveDiT-OTFM      &   10 & 0.0157  & 0.00056 & \textbf{0.830} \\

& WaveDiT-CFM & 10 & \textbf{0.0039} & \underline{0.00010} & \underline{0.834} \\
\midrule
\multirow{2}{*}{\rotatebox[origin=c]{90}{\scriptsize\textit{Abl.}}}

& Voxel-space             &   10 & 0.0045  & \textbf{0.00009} & {0.878} \\

& \textit{w/o}\,Morpheus  &   10 & \multicolumn{1}{c}{0.0295} & \multicolumn{1}{c}{0.00053} & \multicolumn{1}{c}{0.862} \\
\bottomrule
\end{tabular}%
}
\end{minipage}
\end{tabular}
\end{table*}

%% file: TABLES/table_BAP_ROI.tex

\begin{table}[t]
\centering
\caption{Downstream evaluation. BAP: Test MAE in years ($\downarrow$); ROI: 95-region average.
$^\dagger$\!Unconditional model.
Best in \textbf{bold}, second \underline{underlined}.}
\label{tab:clinical}

\setlength{\tabcolsep}{3.5mm} 
\renewcommand{\arraystretch}{0.9} 

\resizebox{\columnwidth}{!}{%
\small
\begin{tabular}{@{}cl r@{\,$\pm$\,}l | r@{\,$\pm$\,}l r@{\,$\pm$\,}l r@{\,$\pm$\,}l@{}}
\toprule
& Method & \multicolumn{2}{c|}{BAP~$\downarrow$} & \multicolumn{2}{c}{iMAE~$\downarrow$} & \multicolumn{2}{c}{KLD~$\downarrow$} & \multicolumn{2}{c@{}}{DICE~$\uparrow$} \\
\midrule
& Real data & 2.92 & 4.08 & \multicolumn{2}{c}{} & \multicolumn{2}{c}{} & \multicolumn{2}{c@{}}{} \\
\midrule
\multirow{9}{*}{\rotatebox[origin=c]{90}{\scriptsize\textit{Baselines}}}
& 3DMD$^\dagger$    & 3.10 & 4.04 & 62.91 & 17.07 & 2.97 & 1.04 & 0.35 & 0.15 \\
& MD$^\dagger$      & 3.72 & 3.24 & 59.33 & 13.12 & 1.45 & 0.55 & 0.05 & 0.10 \\
& WDM$^\dagger$     & 3.93 & 6.17 & 52.65 & 15.33 & 2.36 & 1.30 & 0.36 & 0.15 \\
& WDMa              & 3.08 & 4.62 & 48.57 & 12.23 & 1.06 & 0.45 & 0.33 & 0.15 \\
& MOTFM$^\dagger$   & 3.61 & 5.59 & 61.00 & 15.45 & 1.06 & 0.58 & 0.42 & 0.16 \\
& MOTFMa            & 2.76 & 3.50 & 50.68 &  9.25 & 1.24 & 0.63 & 0.41 & 0.16 \\
& MLDM              & 2.68 & 3.76 & 48.72 & 12.36 & \underline{0.97} & 0.47 & 0.32 & 0.15 \\
& BS                & 3.08 & 3.95 & 48.84 & 12.22 & 1.04 & 0.46 & 0.33 & 0.15 \\
& FlowLet           & 2.72 & 3.85 & \underline{45.18} & 11.11 & 1.18 & 0.56 & \underline{0.44} & 0.16 \\
\midrule
\multirow{3}{*}{\rotatebox[origin=c]{90}{\scriptsize\textit{Ours}}}
& WaveDiT-RFM    & 2.73 & 4.13 & 62.96 & 17.15 & 2.94 & 1.06 & 0.33 & 0.15 \\
& WaveDiT-OTFM   & \underline{2.51} & 3.23 & 46.83 & 11.81 & 1.02 & 0.51 & 0.42 & 0.16 \\
& WaveDiT-CFM    & \textbf{2.44} & 3.19 & \textbf{42.49} & 12.96 & \textbf{0.96} & 0.48 & \textbf{0.46} & 0.16 \\
\midrule
\multirow{2}{*}{\rotatebox[origin=c]{90}{\scriptsize\textit{Abl.}}}
& Voxel-space           & 3.04 & 4.52 & 51.30 & 12.74 & 1.05 & 0.47 & 0.43 & 0.16 \\
& \textit{w/o}\,Morpheus & 2.98 & 4.27 & 54.58 & 13.77 & 1.01 & 0.45 & 0.41 & 0.16 \\
\bottomrule
\end{tabular}%
}
\end{table}

%% file: SECTIONS/4_conclusion.tex
\section{Conclusion}

We introduced WaveDiT, a wavelet-domain conditional flow matching model for 3D brain MRI synthesis that combines an HDiT backbone with the Morpheus state-aware uncertainty scheduler. In our experiments, WaveDiT-CFM achieved strong global distributional scores while also improving brain age prediction and ROI-level anatomical metrics compared to existing diffusion and flow-based baselines under a low-step sampling regime. Ablation studies suggest that both the wavelet representation and state-aware uncertainty weighting contribute to stabilizing training and maintaining region-level anatomical plausibility. This work is limited to T1-weighted MRI and does not include expert reader studies, so we restrict our claims to quantitative proxies of anatomical fidelity and clinical utility. As future directions, we plan to explore WaveDiT on additional 3D imaging domains and modalities, such as CT, and to investigate richer conditioning schemes beyond age.